\documentclass[sigconf]{acmart}
\AtBeginDocument{%
  }

\copyrightyear{2026}
\acmYear{2026}
\setcopyright{cc}
\setcctype{by}
\acmConference[CIKM '26]{Proceedings of the 35th ACM International Conference on Information and Knowledge Management}{November 07--11, 2026}{Rome, Italy}
\acmBooktitle{Proceedings of the 35th ACM International Conference on Information and Knowledge Management (CIKM '26), November 07--11, 2026, Rome, Italy}
\acmDOI{10.1145/3799682.3840696}
\acmISBN{979-8-4007-2539-5/2026/11}

\usepackage{multirow}
\usepackage{amsmath}
\usepackage{comment}
\usepackage{xspace}

\usepackage{graphicx}
\usepackage[table,xcdraw]{xcolor}
\usepackage{diagbox}
\usepackage{algorithm}
\usepackage{algpseudocode}
\usepackage{enumitem}
\usepackage{subfig}

\usepackage[normalem]{ulem}
\useunder{\uline}{\ul}{}

\newcommand{\modelname}{QA-Merging\xspace}

\begin{document}

\title{QA-Merging: Query-Adaptive Reasoning via Layer Selective Model Merging}

\author{Zhaofeng Zhong}
\affiliation{%
  \institution{The University of Queensland}
  \department{School of Electrical Engineering and Computer Science}
  \city{Brisbane}
  \country{Australia}
}
\email{zhaofeng.zhong@uq.edu.au}

\author{Wei Yuan}
\affiliation{%
  \institution{The University of Queensland}
  \department{School of Electrical Engineering and Computer Science}
  \city{Brisbane}
  \country{Australia}
}
\email{warrenyuan97@gmail.com}

\author{Tong Chen}
\affiliation{%
  \institution{The University of Queensland}
  \department{School of Electrical Engineering and Computer Science}
  \city{Brisbane}
  \country{Australia}
}
\email{tong.chen@uq.edu.au}

\author{Liang Qu}
\affiliation{%
  \institution{Edith Cowan University}
  \city{Perth}
  \country{Australia}
}
\email{leoqu94@outlook.com}

\author{Xiangyu Zhao}
\affiliation{%
  \institution{City University of Hong Kong}
  \city{Hong Kong}
  \country{China}}
\email{xianzhao@cityu.edu.hk}

\author{Quoc Viet Hung Nguyen}
\affiliation{%
  \institution{Griffith University}
  \city{Gold Coast}
  \country{Australia}
}
\email{henry.nguyen@griffith.edu.au}

\author{Hongzhi Yin}
\authornote{Corresponding author.}
\affiliation{%
  \institution{The University of Queensland}
  \department{School of Electrical Engineering and Computer Science}
  \city{Brisbane}
  \country{Australia}
}
\email{db.hongzhi@gmail.com}

\renewcommand{\shortauthors}{Zhaofeng Zhong et al.}

\begin{abstract}
Recent large reasoning models (LRMs) have achieved strong performance on complex reasoning tasks by generating a long chain-of-thought (Long-CoT). However, such lengthy reasoning is often unnecessary for simple queries, leading to additional computation and latency. Existing approaches to adaptive reasoning typically rely on retraining the model or designing sophisticated prompting, which are either prohibitively expensive or highly sensitive to the prompt formulation. Model merging provides a more balanced alternative for adaptive reasoning by avoiding expensive training and integrating Long-CoT and Short-CoT behaviors. However, existing merging methods are often static and input-agnostic, or rely on costly all-layer calibration, which limits their effectiveness for query-adaptive reasoning. To tackle these challenges, we propose Query-adaptive Layer Selective Merging (QA-Merging), an activation-based merging framework that integrates a Long-CoT model and a Short-CoT model to obtain a query-adaptive reasoner without training from scratch or requiring large-scale additional data. QA-Merging first constructs a small pattern-labeled calibration set that assigns each query an appropriate reasoning pattern. Motivated by our empirical analysis that Long-CoT and Short-CoT behaviors diverge unevenly across transformer layers, QA-Merging identifies layers with high reasoning pattern divergence and calibrates only these layers through feature alignment and contrastive shaping, while applying closed-form hidden-state correction to the remaining layers. Experiments on seven widely used reasoning benchmarks across two model scales demonstrate that QA-Merging reduces inference cost and maintains strong performance. The code is available at https://github.com/Zephyr-Zhong/QA-Merging.
\end{abstract}

%
%


%

\maketitle

\section{Introduction}
Large reasoning models (LRMs) have recently achieved strong performance on complex reasoning tasks, ranging from mathematical problem solving~\citep{SolveReasonLLM-nips22-Aitor,plumPromptLearning-pan-etal-2024,deepseek-math-2024-CoRR} and logical deduction~\citep{CoTPromptLLM-2022-WeiJason} to agentic assistants~\citep{AgenticReasoning-2025-Wu,ReSearch-reasonWithSerachRL-chen2025}. 
A key driver is their ability to generate long chains of thought (Long-CoT), in which the model iteratively self-assesses, mitigates errors, and verifies intermediate steps before producing a final answer~\citep{longgenbench-liu-etal-2024,distilling2to1-yu2025,FromSystem1to2Survey-li2025}. 
However, although Long-CoT is beneficial for difficult problems, it can be counterproductive on simple tasks that require few reasoning steps: models may ``overthink'' by introducing unnecessary intermediate reasoning~\citep{simpleTestTimeScale-muennighoff-etal-2025-s1,surveytesttimescalinglarge-zhang2025}. 
This overthinking not only increases inference cost~\citep{missingpremiseexacerbatesoverthinking-fan2025,DoNotOverthink-chen2025,DangerOverthinkingExaminingReasoning-cuadron2025}, but can also hurt accuracy by amplifying the chance of spurious reasoning and obscuring the straightforward solution~\citep{DifficultyAdaptive-shen-etal-2025}.

Many recent studies have explored how to mitigate the inefficiency and overthinking of LRMs~\citep{Efficient-Reasoning-Survey-feng2025,ReasoningOfSmallLMsrivastava2025,sui2025stopoverthinkingsurveyefficient}. 
A prominent direction is adaptive reasoning, where a model dynamically selects between Long-CoT and Short-CoT modes depending on the estimated problem complexity~\citep{adaptthink-zhang-etal-2025,alomrani2025reasoningbudgetsurveyadaptive}. 
Broadly, existing adaptive reasoning approaches fall into two paradigms. 
Training-free methods, often implemented through prompt-guided strategies~\citep{xu2025-CoD-ChainofDraft-prompt}, can introduce adaptivity without additional training cost, yet they rely heavily on instruction following and can be sensitive to prompt constraints and query phrasing~\citep{LRM-promptsave-zhu2025,zhu2025-Survey-AdaptiveThinkInLRM}. 
Training-based methods optimize LRMs to exhibit adaptive thought processes via supervised fine-tuning (SFT)~\citep{yu2025longshortCoTMixSFT,CoTValue-ma-etal-2025} or reinforcement learning (RL)~\citep{fang2025thinkless,ARM-wu2025}, but they typically require large-scale training data and incur significant training cost. 
As a result, achieving adaptive reasoning that is both efficient and effective remains non-trivial.

Thus, model merging techniques have emerged as a more balanced route toward adaptive reasoning.
Instead of training a model from scratch or performing end-to-end fine-tuning on a large-scale training dataset, model merging combines the parameters of multiple task- or style-specific models into a single model~\citep{modelmerginginLlmsMllms-Yang2024}. 
This paradigm is particularly suitable for adaptive reasoning, where Long-CoT and Short-CoT behaviors are often learned by separate models. 
By merging these models, we can obtain a unified model that inherits the deep reasoning capabilities of the Long-CoT model and the concise generation behavior of the Short-CoT model~\citep{wu2025unlockingefficientL2S,AIM-merging-nobari2025,ACM-merging-neurips25}, while avoiding the additional computational overhead of training-based adaptation.
As a result, compared with prompt- or training-based solutions, model merging appears to be a superior alternative for efficient and effective adaptive reasoning.

Despite the promises, existing merging methods are not directly tailored to query-adaptive reasoning, with arithmetic merging being the most straightforward yet static method.
Such methods directly manipulate model parameters through simple arithmetic operations, such as average interpolation~\citep{ModelSoups-averageWeight-2022Kamalika}, task vector addition~\citep{editing-TAMerge-ilharco2023}, or sign conflict resolution~\citep{TIES-merging-NEURIPS2023}. 
Though they operate directly in the parameter space without requiring additional optimization, their merging rules are typically static and input-agnostic, leading to coarse-grained parameter ensemble with suboptimal performance in downstream tasks~\citep{liu2025sens-merging}.
To obtain finer-grained control, activation-based merging methods construct calibration sets and utilize activations~\citep{AIM-merging-nobari2025} or gradients~\citep{liu2025sens-merging} to optimize layer-wise merging coefficients.
By incorporating behavioral signals from the candidate models, they usually provide more nuanced and performant merging than arithmetic operations. 
Nevertheless, such calibration often requires activation or gradient computation across all layers, where the additional computation overhead offsets the efficiency advantage of model merging.
Furthermore, in the context of adaptive reasoning, another limitation of existing model merging methods is that they are commonly optimized toward the goal of uniformly reducing response length regardless of the query difficulty.
This global compression is not query-adaptive. Ideally, when removing redundant reasoning for simple problems, detailed derivations required for correctly answering difficult queries should be encouraged rather than suppressed.

To address these challenges, we propose \textbf{Query-adaptive Layer Selective Merging (\modelname)}, an activation-based merging framework for query-adaptive reasoning. Instead of mandating time-consuming optimizations across all layers, QA-Merging can selectively calibrate a subset of layers while enabling query-level selection between Long-CoT and Short-CoT reasoning patterns. Our design is motivated by an empirical investigation (presented in Section~\ref{sec:observation}) that Long-CoT and Short-CoT behaviors do not diverge uniformly across transformer layers.
By measuring the layer-wise hidden-state distance between the Long-CoT and Short-CoT models, a small subset of middle-to-late layers are found to be the major contributor to their  diverging reasoning patterns, while many other layers exhibit relatively similar representations.
Based on this observation, \modelname first constructs a compact pattern-labeled calibration set that determines the preferred reasoning pattern (Long-CoT vs.\ Short-CoT) for each query. 
We then perform layer-wise merging. 
For high-divergence layers, \modelname calibrates element-wise merging coefficients through feature alignment and contrastive shaping, encouraging the merged model to match the query-preferred reasoning pattern. For the remaining low-divergence layers, a closed-form solution based on feature alignment is designed to efficiently calibrate their generated hidden-states.
This design enables \modelname to preserve the fine-grained behavioral control of activation-based merging while avoiding unnecessary all-layer calibration, enabling the merged model to generate detailed reasoning for difficult queries and concise reasoning for simpler ones.

We evaluate \modelname on seven widely used reasoning benchmarks across two model scales. 
Experimental results show that \modelname achieves a favorable accuracy--efficiency trade-off compared with prompt-guided and training-based adaptive reasoning approaches, and both arithmetic and activation-based model merging baselines. 
On the Qwen2.5-1.5B model pair, the results show that \modelname even improves accuracy over the Long-CoT base model while reducing response length by 70\%. 
This demonstrates that query-adaptive model merging is an effective and efficient alternative to costly training methods for adaptive reasoning.

Our contributions are summarized as follows:
\begin{itemize}[leftmargin=*]
 \item We propose a model merging framework, \modelname, for \emph{query-level adaptive reasoning}, enabling a single merged model to dynamically balance the deep reasoning capability of Long-CoT with the efficiency of Short-CoT generation. Specifically, we first develop a novel training data construction strategy to build a compact yet effective pattern-labeled calibration dataset, and then incorporate contrastive shaping for feature alignment, guiding the merged model toward query-adaptive reasoning.
    \item To further improve the calibration efficiency of \modelname, we propose a layer-selective strategy based on reasoning pattern divergence, which optimizes only the most important layers and applies closed-form hidden-state correction to the remaining layers to balance model performance with merging efficiency.
    \item We conduct extensive experiments on seven widely used reasoning benchmark datasets. The results show that \modelname achieves a more desirable accuracy-efficiency trade-off than existing adaptive reasoning baselines and merging approaches, attaining near Long-CoT base model performance while significantly reducing generated tokens.
\end{itemize}

\section{Preliminary}
\begin{figure}[]
    \centering
    \includegraphics[width=\columnwidth]{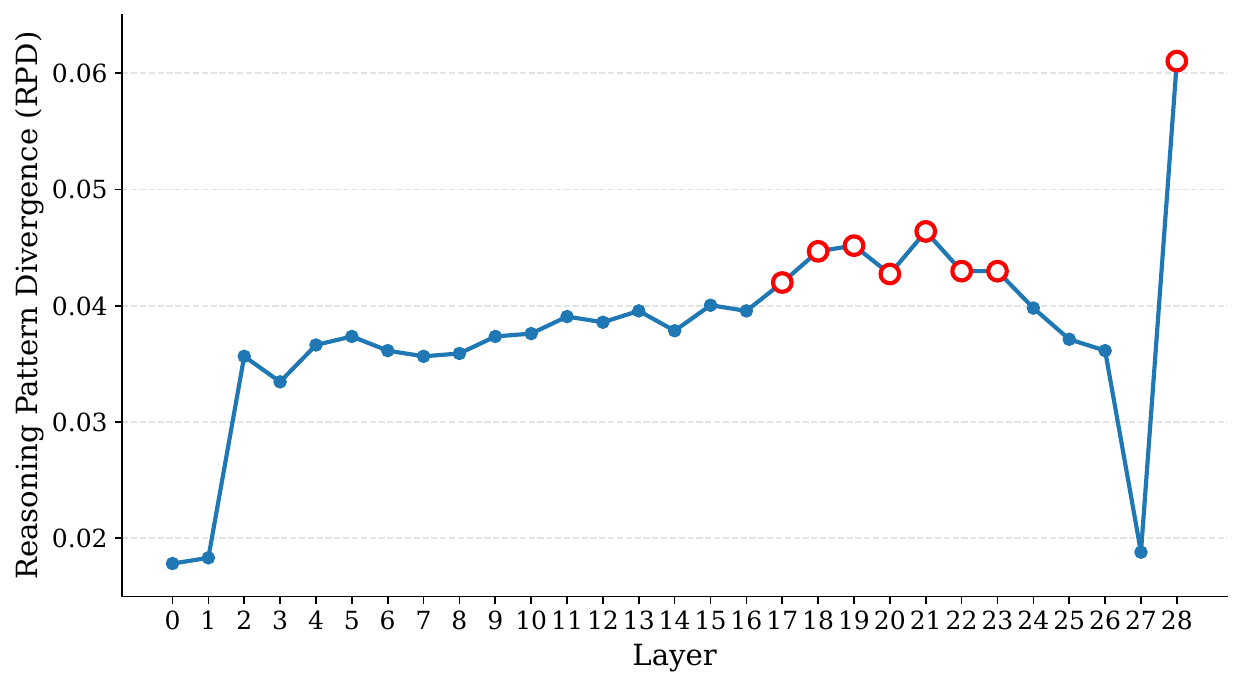}
    \caption{Layer-wise reasoning-pattern divergence (RPD) between $\theta_L$ and $\theta_S$.
    The top 8 (30\%) layers are highlighted in red.
    }
    \label{fig:layer_divergence}
\end{figure}

\begin{figure*}[ht]
    \centering
    \includegraphics[width=\textwidth]{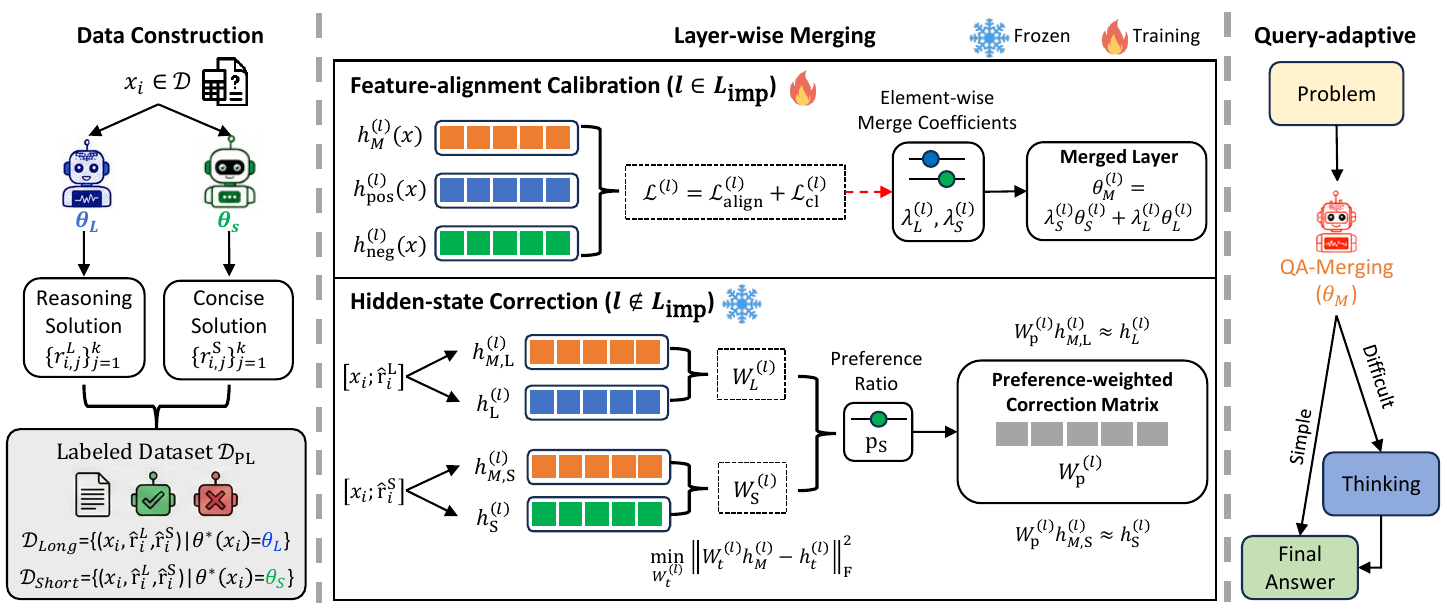}
    \caption{An overview of \modelname. It consists of two key components: (1) Constructing a pattern-labeled (PL) dataset that identifies the optimal reasoning pattern (Long-CoT or Short-CoT) and the shortest correct responses $\hat{r}$ for each query $x$; (2) Performing layer-wise merging via feature alignment calibration for important layers, and hidden-state correction for remaining layers. See Sec.~\ref{sec:methodology} for details.}
    \label{fig:overview}
\end{figure*}

\subsection{Model Merging}
Model merging aims to combine $T$ task-specific models \(\{\theta_i\}_{i=1}^{T}\), typically specialized for different tasks or behaviors.
The goal is to obtain a single parameter set, \(\theta_M\), without additional end-to-end retraining and preserving the strengths of the constituent models. A simple and widely used approach is linear weight merging:
\begin{equation}
    \theta_M = \sum_{i=1}^{T} \lambda_i \theta_i,
\end{equation}
where \(\lambda_i\) controls the contribution of model \(\theta_i\) (often with \(\sum_i \lambda_i = 1\)). 
However, because the coefficients are usually pre-defined, naive linear merging can introduce representation bias and yield suboptimal performance~\citep{surgeryRepresentation-Yang-ICML2024}. 
To alleviate this issue, post-calibration merging methods have been proposed~\citep{dai2025-leverageLinearity,prodistill_merge_xu2025scalable,WuAAAi26-react}, which make the merging coefficients learnable and calibrate the merged model toward task-specific optima, thereby reducing knowledge loss.

\subsection{Empirical Analysis}
\label{sec:observation}
Simply calibrating merging coefficients across all layers can introduce unnecessary computational overhead, especially when reasoning-pattern divergence (RPD) is not evenly distributed throughout the model.
To investigate this phenomenon, we compare the hidden representations of Long-CoT (DeepSeek-R1-Distill-Qwen-1.5B) and a Short-CoT (Qwen2.5-Math-1.5B) on a mixed reasoning corpus $\mathcal{D}_{mix}$, constructed from GSM8K, MATH500, and AIME24 (dataset details in Section~\ref{sec:exp_setup}).
For each input query $x \in \mathcal{D}_{mix}$, let $h_L^{(l)}(x)$ and $h_S^{(l)}(x)$ denote the hidden representations produced by $\theta_L$ and $\theta_S$ at layer $l$, respectively.
We quantify the layer-wise discrepancy between the two reasoning patterns using cosine distance:
\begin{equation}
D^{(l)}(x)=1 -
\frac{
h^{(l)}_L(x) \cdot h^{(l)}_S(x)
}{
\|h^{(l)}_L(x)\|_2 \, \|h^{(l)}_S(x)\|_2
}.
\label{eq:cos_dist}
\end{equation}
We then average this quantity over $\mathcal{D}_{mix}$ to obtain the RPD for layer $l$:
\begin{equation}
\mathrm{RPD}^{(l)}=
\frac{1}{|\mathcal{D}_{mix}|}
\sum_{x \in \mathcal{D}_{mix}}
D^{(l)}(x).
\label{eq:RPD}
\end{equation}
A larger $\mathrm{RPD}^{(l)}$ indicates that the Long-CoT and Short-CoT models induce more separated intermediate representations at layer $l$, suggesting that this layer is more strongly associated with reasoning-pattern differentiation.
Conversely, a smaller $\mathrm{RPD}^{(l)}$ indicates that the two models produce more similar representations at that layer.

As shown in Figure~\ref{fig:layer_divergence}, the divergence between Long-CoT and Short-CoT models is highly non-uniform across layers.
The discrepancy is concentrated in several middle-to-late layers, where the RPD values are higher than those of earlier layers.
In addition, we compute the variance of RPD values across all layers. 
The mean RPD is $0.038$, and the across-layer variance is $6.92 \times 10^{-5}$. This corresponds to a coefficient of variation (CV) of $22.0\%$, where CV measures the relative variability of RPD values with respect to the mean across layers. 
Moreover, the top-8 layers have an average RPD of $0.046$, which is $1.3$ times that of the remaining layers ($0.035$). 
This empirical analysis indicates that reasoning pattern discrepancy is not evenly distributed across all layers. Therefore, \modelname ranks layers by RPD and calibrates only the top-$K$ high-divergence layers, while applying hidden-state correction to the remaining layers.

\subsection{Problem Formulation}
Given a query prompt $x$, an LLM parameterized by $\theta$ generates a corresponding solution $y=[y_{1}, ..., y_{n}]$ by sampling from the conditional distribution $\pi_{\theta}(\cdot|x)$. This distribution factorizes auto-regressively as:
\begin{equation}
    \pi_{\theta}(y|x) = \prod_{i=1}^{n} \pi_{\theta}(y_{i} \mid x, y_{<i}).
    \label{eq:prob_decomp}
\end{equation}

To improve the solution quality on reasoning tasks, Chain-of-Thought (CoT) prompting has been widely adopted, as it encourages models to produce intermediate steps that facilitate self-evaluation and verification. 
According to the granularity of the generated steps, CoT reasoning can be classified into Long-CoT and Short-CoT~\citep{Ada-R1-neurips25-luo2025dpoMerge}. 
Long-CoT yields detailed, reflective reasoning that benefits complex queries but increases generation length and latency. 
Short-CoT, in contrast, produces concise reasoning (or even direct answers) with lower cost, but can struggle on harder problems that require multi-step deliberation.

In this paper, we consider two LLMs: a Long-CoT model $\theta_L$ and a Short-CoT model $\theta_{S}$. 
Our goal is to construct a merged model $\theta_M$ that adapts its reasoning pattern to each query, producing Long-CoT when deeper reasoning is necessary and Short-CoT when a concise response suffices.
Following prior work~\citep{AIM-merging-nobari2025, ACM-merging-neurips25} on model merging for efficient reasoning, we evaluate our method only in the homogeneous setting.

\section{Methodology}
\label{sec:methodology}
In this paper, we propose \modelname, a layer selective model merging framework for query-adaptive reasoning. 
We aim to obtain a single merged model that preserves the deep reasoning capability of $\theta_L$ on hard queries and retains the concise generation behavior of $\theta_S$ on simple queries. 
Rather than optimizing all transformer layers, \modelname adopts a layer-selective strategy. 
The layers with high RPD are merged by feature-alignment calibration and contrastive shaping, while the remaining layers are corrected by closed-form hidden-state correction. 
The overall framework is shown in Fig.~\ref{fig:overview}.
We present the technical details of each component below.

\begin{algorithm}[t]
\caption{Layer-wise Calibration and Correction}
\label{alg:pmm}
\begin{algorithmic}[1]
\Require Long-CoT model $\theta_L$; Short-CoT model $\theta_S$; pattern-labeled dataset $\mathcal{D}_{PL}$; important layers $L_{\mathrm{imp}}$; preference ratio $p_S$.
\Ensure Final merged model $\theta_{\mathrm{final}}$

\For{$l=1,\ldots,L$}
    \If{$l\in L_{\mathrm{imp}}$}
        \State Initialize element-wise coefficients $\lambda_L^{(l)}$ and $\lambda_S^{(l)}$
        \State Initialize $\theta_M^{(l)}$ with Eq.~\eqref{eq:merged_layer_param}
        \Repeat
            \State Sample a mini-batch from $\mathcal{D}_{\mathrm{PL}}$
            \State Extract layer features $h_M^{(l)}$, $h_{\mathrm{pos}}^{(l)}$, $h_{\mathrm{neg}}^{(l)}$ with Eq.~\eqref{eq:features}
            \State Compute $\mathcal{L}_{\mathrm{align}}^{(l)}$ and $\mathcal{L}_{\mathrm{cl}}^{(l)}$ with Eq.~\eqref{eq:mse_loss} and Eq.~\eqref{eq:cl}
            \State Update $\lambda_L^{(l)}$ and $\lambda_S^{(l)}$ to minimize $\mathcal{L}^{(l)}$ with Eq.~\eqref{eq:joint_selected_layers}
        \Until{convergence}
    \Else
        \State Initialize $\theta_M^{(l)}$ with average merging
        \State Construct samples with responses $s_i^t$ with Eq.~\eqref{eq:sample_with_response}
        \State Extract hidden states from $\theta_M$, $\theta_L$ and $\theta_S$
        \State Compute pattern-specific $W_t^{(l)}$ with Eq.~\eqref{eq:task-wise-matrix}
        \State Compute the preference-weighted $W_p^{(l)}$ with Eq.~\eqref{eq:layerwise_react_preference}
    \EndIf
\EndFor

\State \Return $\theta_{\mathrm{final}}=
\left(
\{\theta_M^{(l)}\}_{l\in L_{\mathrm{imp}}},
\{\theta_M^{(l)},W_p^{(l)}\}_{l\notin L_{\mathrm{imp}}}
\right)$
\end{algorithmic}
\end{algorithm}

\subsection{Pattern-Labeled Data Construction}
A key challenge in adaptive reasoning is to determine which reasoning pattern is preferred for each query. 
To address this problem, we construct a small pattern-labeled calibration dataset by comparing the empirical correctness and generation efficiency of the Long-CoT and Short-CoT models on the same seed queries.
Given a seed dataset
\begin{equation}
\mathcal{D}=\{(x_i,y_i)\}_{i=1}^{N},
\end{equation}
where $x_i$ is the query and $y_i$ is the ground-truth answer.
For each query $x_i$, we sample $k$ responses from both models, denoted by $\{r_{i,j}^{L}\}_{j=1}^{k}$ and $\{r_{i,j}^{S}\}_{j=1}^{k}$, respectively.
We estimate its empirical correctness on query $x_i$ as
\begin{equation}
\mathbb{E}(\theta,x_i)
=
\frac{1}{k}
\sum_{j=1}^{k}
\mathbb{I}\left[r_{i,j}^{\theta}=y_i\right],
\qquad
\theta\in\{\theta_L,\theta_S\},
\label{eq:expected_acc}
\end{equation}
and its average response length as
\begin{equation}
\mathbb{T}(\theta,x_i)
=
\frac{1}{k}
\sum_{j=1}^{k}
\mathrm{len}\left(r_{i,j}^{\theta}\right),
\qquad
\theta\in\{\theta_L,\theta_S\},
\label{eq:expected_len}
\end{equation}
where $\mathbb{I}[\cdot]$ is the indicator function and $\mathrm{len}(\cdot)$ denotes the number of generated tokens.

We aim to preserve \(\theta_S\)'s generation efficiency for queries that both models answer correctly.
Thus, we use both correctness and response length to divide the seed dataset \(\mathcal{D}\) into two subsets, namely \(\mathcal{D}_{\mathrm{Long}}\) and \(\mathcal{D}_{\mathrm{Short}}\), corresponding to hard queries and simple queries, respectively. 
We determine the preferred reasoning pattern for each query as follows:
\begin{equation}
\theta^{*}(x_i)=
\begin{cases}
\theta_L, & \mathbb{E}(\theta_L,x_i)>\mathbb{E}(\theta_S,x_i),\\[3pt]
\theta_S, & \mathbb{E}(\theta_S,x_i)>\mathbb{E}(\theta_L,x_i),\\[3pt]
\arg\min\limits_{\theta\in\{\theta_L,\theta_S\}}\mathbb{T}(\theta,x_i), & \mathbb{E}(\theta_L,x_i)=\mathbb{E}(\theta_S,x_i).
\end{cases}
\label{eq:assignment_rule}
\end{equation}

In addition, we retain representative correct responses from both reasoning models. For each query $x_i$, we define the set of correct Long-CoT responses and correct Short-CoT responses as
\begin{equation}
    \mathcal{C}^{L}_{i}=
    \left\{
        r^{L}_{i,j}
        \mid
        \mathbb{I}\left[r_{i,j}^{L}=y_i\right]
    \right\},
    \quad
    \mathcal{C}^{S}_{i}=
    \left\{
        r^{S}_{i,j}
        \mid
        \mathbb{I}\left[r_{i,j}^{S}=y_i\right]
    \right\}.
\end{equation}
For each query, we select the shortest correct response from each model:
\begin{equation}
\begin{aligned}
    \hat{r}^{L}_{i}
    =
    \arg\min_{r \in \mathcal{C}^{L}_{i}}
    \operatorname{len}(r),
    \quad
    \hat{r}^{S}_{i}
    =
    \arg\min_{r \in \mathcal{C}^{S}_{i}}
    \operatorname{len}(r).
\label{short_correct_responses}
\end{aligned}
\end{equation}

Accordingly, $\mathcal{D}$ is partitioned into Long-CoT preferred and Short-CoT preferred subsets:
\begin{equation}
\begin{aligned}
    \mathcal{D}_{\mathrm{Long}}=
    \left\{
    \left(
        x_i,
        \hat{r}^{L}_{i},
        \hat{r}^{S}_{i}
    \right)
    \mid
    \theta^{*}(x_i)=\theta_L
    \right\}, \\
    \mathcal{D}_{\mathrm{Short}}=
    \left\{
    \left(
        x_i,
        \hat{r}^{L}_{i},
        \hat{r}^{S}_{i}
    \right)
    \mid
    \theta^{*}(x_i)=\theta_S
    \right\}.
\end{aligned}
\end{equation}
The final pattern-labeled dataset is defined as:
\begin{equation}
    \mathcal{D}_{\mathrm{PL}}
    =
    \mathcal{D}_{\mathrm{Long}}
    \cup
    \mathcal{D}_{\mathrm{Short}}.
\end{equation}

This construction provides query-level supervision for query-adaptive merging, enabling the merged model to preserve the stronger reasoning ability of \(\theta_L\) on hard queries and retain the efficiency of \(\theta_S\) on simpler ones.

\subsection{Layer-selective Strategy}
The Long-CoT and Short-CoT models do not differ uniformly across all layers. 
Following the layer-wise RPD analysis, we compute the $\mathrm{RPD}^{(l)}$ as defined in Section~2.2. Layers with larger $\mathrm{RPD}$ show stronger reasoning pattern discrepancy.
This observation suggests that uniformly optimizing all layers is unnecessary: layers with high divergence should be selected for subsequent merging calibration, while layers with low divergence can be solved by a hidden-state correction.

We rank all $L$ layers by $\mathrm{RPD}^{(l)}$ in descending order and select the top $K$ layers as the important layer set:
\begin{equation}
{L}_{\mathrm{imp}}
=
\mathrm{TopK}\left(\{\mathrm{RPD}^{(l)}\}_{l=1}^{L},K\right).
\label{eq:important_layer_set}
\end{equation}
For important layers $l\in{L}_{\mathrm{imp}}$, the element-wise merging coefficients are learnable and calibrated using the feature-alignment objective described in Section~\ref{sec:fa_calibrate}. 
For the remaining layers $l\notin{L}_{\mathrm{imp}}$, we do not calibrate their merging coefficients to avoid computation cost. 
Instead, these layers are initialized by average parameter merging and handled later by closed-form hidden-state correction in Section~\ref{sec:representation_correction}.

\subsection{Feature-Alignment Calibration with Contrastive Shaping}
~\label{sec:fa_calibrate}
A naive linear interpolation between \(\theta_L\) and \(\theta_S\) may blur their reasoning behaviors and therefore fail to perform the desired query-adaptive reasoning behavior. 
We formulate model merging as a layer-wise hidden-state alignment problem to learn element-wise merging coefficients on the constructed \(\mathcal{D}_{\mathrm{PL}}\).

Based on \(\mathcal{D}_{\mathrm{Long}}\) and \(\mathcal{D}_{\mathrm{Short}}\), we define the positive and negative models for each query \(x\) as
\begin{equation}
\label{eq:posneg_def}
(\theta_{\mathrm{pos}}(x), \theta_{\mathrm{neg}}(x)) =
\begin{cases}
(\theta_L, \theta_S), & x \in \mathcal{D}_{\mathrm{Long}},\\
(\theta_S, \theta_L), & x \in \mathcal{D}_{\mathrm{Short}}.
\end{cases}
\end{equation}
For hard queries in \(\mathcal{D}_{\mathrm{Long}}\), \(\theta_L\) serves as the positive model and \(\theta_S\) serves as the negative model. For simple queries in \(\mathcal{D}_{\mathrm{Short}}\), the assignment is reversed.
For each selected layer \(l \in L_{\mathrm{imp}}\), let \(\lambda_L^{(l)}\) and \(\lambda_S^{(l)}\) denote the element-wise merging coefficients, and the merged parameters at layer \(l\) are given by
\begin{equation}
\label{eq:merged_layer_param}
\theta_M^{(l)} = \lambda_S^{(l)} \odot\theta_S^{(l)} + \lambda_L^{(l)} \odot \theta_L^{(l)}.
\end{equation}

For a given input \(x\), we define the internal hidden-state at layer $l$ of model $\theta$ as $h_M^{(l)}(x) \in \mathbb{R}^{|x| \times d}$, where $|x|$ is the number of input tokens and $d$ is the hidden dimension.
Let \(h_M^{(l-1)}(x)\), \(h_{\mathrm{pos}}^{(l-1)}(x)\), and \(h_{\mathrm{neg}}^{(l-1)}(x)\) denote the inputs to layer \(l\) in $\theta_M$, \(\theta_{\mathrm{pos}}\), and \(\theta_{\mathrm{neg}}\), respectively.
 Let \(\varphi^{(l)}(\cdot)\) denote the feature mapping of layer \(l\). Then the corresponding hidden states are computed by
\begin{equation}
\label{eq:features}
\begin{aligned}
h_M^{(l)}(x) &= \varphi^{(l)}\!\left(\theta_M^{(l)},\, h_M^{(l-1)}(x)\right),\\
h_{\mathrm{pos}}^{(l)}(x) &= \varphi^{(l)}\!\left(\theta_{\mathrm{pos}}^{(l)},\, h_{\mathrm{pos}}^{(l-1)}(x)\right),\\
h_{\mathrm{neg}}^{(l)}(x) &= \varphi^{(l)}\!\left(\theta_{\mathrm{neg}}^{(l)},\, h_{\mathrm{neg}}^{(l-1)}(x)\right).
\end{aligned}
\end{equation}

We first align the merged model with the query-specific positive model by minimizing the squared \(\ell_2\) distance:
\begin{equation}
\label{eq:mse_loss}
\mathcal{L}_{\mathrm{align}}^{(l)} = \frac{1}{|x|}
\left\|
h_M^{(l)}(x) - h_{\mathrm{pos}}^{(l)}(x)
\right\|^2.
\end{equation}

By optimizing \(\{\lambda_L^{(l)}, \lambda_S^{(l)}\}\) under \(\mathcal{L}_{\mathrm{align}}^{(l)}\), \modelname encourages the merged model to inherit the internal representation of the model preferred by query \(x\).

To better separate these two reasoning patterns, we introduce a contrastive objective that pulls the merged model's feature closer to the positive model's and pushes it away from the negative model's. We define a binary contrastive loss:
\begin{equation}
\mathcal{L}_{\mathrm{cl}}^{(l)} =
-\log
\frac{
\exp\left(h_M^{(l)\top}(x) h_{\mathrm{pos}}^{(l)}(x) / \tau\right)
}{
\sum_{a \in \{\mathrm{pos}, \mathrm{neg}\}}
\exp\left(h_M^{(l)\top}(x) h_{a}^{(l)}(x) / \tau\right)
},
\label{eq:cl}
\end{equation}
where \(\tau\) is a temperature to control the strength of the contrastive term.

Finally, for each layer \(l\), we optimize a joint objective that combines feature alignment with contrastive shaping:
\begin{equation}
\mathcal{L}^{(l)} =
\mathcal{L}_{\mathrm{align}}^{(l)} +
\omega \cdot \mathcal{L}_{\mathrm{cl}}^{(l)},
\qquad
l\in{L}_{\mathrm{imp}},
\label{eq:joint_selected_layers}
\end{equation}
where \(\omega\) controls the strength of the contrastive term.

\subsection{Closed-form Hidden-state Correction}
\label{sec:representation_correction}

For the remaining layers $l\notin{L}_{\mathrm{imp}}$, we do not introduce additional learnable merging coefficients. 
Although these layers show lower RPD than the important layers, the average merged model can still suffer from hidden representation distortion relative to the Long-CoT and Short-CoT models. 
We treat this distortion as a linear transformation in hidden-state space and compute a closed-form correction matrix for each remaining layer.

Specifically, we estimate a transformation matrix that maps the hidden representations of the initialized merged model toward the corresponding representation space. 
For each reasoning pattern, we construct a sample
\begin{equation}
    s_i^{t} = [x_i;\hat{r}^{t}_{i}],
    \qquad
    t\in\{L,S\},
\label{eq:sample_with_response}
\end{equation}
where $[\,;\,]$ denotes text concatenation. 
For each layer $l$, we use a set of constructed samples $\{s_i^{t}\}$ to extract the hidden states from the merged model, $h_{M,t}^{(l)}(s_i^{t})$,  and those from the specialist model, $h_{L}^{(l)}(s_i^{t})$ and $h_{S}^{(l)}(s_i^{t})$.
For each reasoning pattern, we seek a linear correction matrix $W_{L}^{(l)}$ and $W_{S}^{(l)}$ such that the corrected merged hidden states approximate the specialist hidden states:
\begin{equation}
\begin{aligned}
    W_{L}^{(l)} h_{M,t}^{(l)}(s_i^{t}) \approx h_{L}^{(l)}(s_i^{t}), \\
    W_{S}^{(l)} h_{M,t}^{(l)}(s_i^{t}) \approx h_{S}^{(l)}(s_i^{t}).
\end{aligned}
\end{equation}
To find the optimal correction matrix, the objective is to minimize the squared Frobenius norm between the corrected merged representations and the target specialist representations:
\begin{equation}
    \min_{W_{t}^{(l)}}
    \left\|
    W_{t}^{(l)}h_{M,t}^{(l)}(s_i^{t}) - h_{t}^{(l)}(s_i^{t})
    \right\|_{F}^{2}.
    \label{eq:task-wise-matrix}
\end{equation}

Thus, for each remaining layer, we obtain two correction matrices for each reasoning pattern, ${W}_{L}^{(l)}$ and ${W}_{S}^{(l)}$.
To enable controllable reasoning behavior, we combine the two pattern-specific corrections using a preference ratio for Short-CoT, $p_S$.
The preference-weighted correction matrix for layer $l$ is computed as
 \begin{equation}
    W_{p}^{(l)}= 
    (1-p_S){W}_{L}^{(l)}+
    p_S{W}_{S}^{(l)}.
    \label{eq:layerwise_react_preference}
\end{equation}

In general, the remaining layers are initialized by average merging and then corrected in the hidden-state space. 
Unlike the important layers, which are calibrated through coefficient optimization, these layers require no gradient-based computation. 
This correction is computed once from response-token hidden states and can be adjusted by the preference ratio $p_S$ to control the Long-CoT and Short-CoT trade-off.
The complete lightweight merging procedure is summarized in Algorithm~\ref{alg:pmm}.

\section{Experiments}

\begin{table*}[pt!]
\caption{Evaluations of different methods on Qwen2.5-1.5B series models. The number in () represents the average response length on the benchmark. The best performance is highlighted in bold.}
\resizebox{0.9\textwidth}{!}{%
\begin{tabular}{ccccccccc}
\hline
\textbf{\diagbox{Model}{Bench}} & {\color[HTML]{1F2328} \textbf{GSM8K}} & {\color[HTML]{1F2328} \textbf{MATH500}} & {\color[HTML]{1F2328} \textbf{\begin{tabular}[c]{@{}c@{}}Minerva \\      Math\end{tabular}}} & \textbf{\begin{tabular}[c]{@{}c@{}}Olympiad\\      Bench\end{tabular}} & {\color[HTML]{1F2328} \textbf{AIME24}} & {\color[HTML]{1F2328} \textbf{AIME25}} & {\color[HTML]{1F2328} \textbf{GPQA}} & \textbf{Avg.} \\ \hline
 & 79.0 & 80.6 & 30.7 & 30.4 & 29.2 & 25.6 & 34.5 & 44.3 \\
\multirow{-2}{*}{\begin{tabular}[c]{@{}c@{}}DeepSeek-R1-1.5B\\ (Long-CoT)\end{tabular}} & \cellcolor[HTML]{DDEBF7}(978) & \cellcolor[HTML]{DDEBF7}(3741) & \cellcolor[HTML]{DDEBF7}(4948) & \cellcolor[HTML]{DDEBF7}(7389) & \cellcolor[HTML]{DDEBF7}(17465) & \cellcolor[HTML]{DDEBF7}(13118) & \cellcolor[HTML]{DDEBF7}(9492) & \cellcolor[HTML]{DDEBF7}(8162) \\ \hline
 & 75.9 & 36.2 & 11.4 & 22.8 & 0.0 & 1.1 & 19.7 & 23.9 \\
\multirow{-2}{*}{\begin{tabular}[c]{@{}c@{}}Qwen2.5-Math-1.5B\\ (Short-CoT)\end{tabular}} & \cellcolor[HTML]{DDEBF7}(118) & \cellcolor[HTML]{DDEBF7}(411) & \cellcolor[HTML]{DDEBF7}(1037) & \cellcolor[HTML]{DDEBF7}(608) & \cellcolor[HTML]{DDEBF7}(865) & \cellcolor[HTML]{DDEBF7}(1119) & \cellcolor[HTML]{DDEBF7}(743) & \cellcolor[HTML]{DDEBF7}(700) \\ \hline
\rowcolor[HTML]{F2F2F2} 
\textit{Prompt-guided} & \multicolumn{1}{l}{\cellcolor[HTML]{F2F2F2}\textit{}} & \multicolumn{1}{l}{\cellcolor[HTML]{F2F2F2}\textit{}} & \multicolumn{1}{l}{\cellcolor[HTML]{F2F2F2}\textit{}} & \multicolumn{1}{l}{\cellcolor[HTML]{F2F2F2}\textit{}} & \multicolumn{1}{l}{\cellcolor[HTML]{F2F2F2}\textit{}} & \multicolumn{1}{l}{\cellcolor[HTML]{F2F2F2}\textit{}} & \multicolumn{1}{l}{\cellcolor[HTML]{F2F2F2}\textit{}} & \multicolumn{1}{l}{\cellcolor[HTML]{F2F2F2}\textit{}} \\
 & 82.9 & 74.2 & 34.2 & 33.0 & 20.0 & 16.7 & 25.8 & 41.0 \\
\multirow{-2}{*}{CoD} & \cellcolor[HTML]{DDEBF7}(1149) & \cellcolor[HTML]{DDEBF7}(2308) & \cellcolor[HTML]{DDEBF7}(2739) & \cellcolor[HTML]{DDEBF7}(3321) & \cellcolor[HTML]{DDEBF7}(3920) & \cellcolor[HTML]{DDEBF7}(3926) & \cellcolor[HTML]{DDEBF7}(3364) & \cellcolor[HTML]{DDEBF7}(2961) \\ \hline
\rowcolor[HTML]{F2F2F2} 
\textit{Training-based} & \multicolumn{1}{l}{\cellcolor[HTML]{F2F2F2}\textit{}} & \multicolumn{1}{l}{\cellcolor[HTML]{F2F2F2}\textit{}} & \multicolumn{1}{l}{\cellcolor[HTML]{F2F2F2}\textit{}} & \multicolumn{1}{l}{\cellcolor[HTML]{F2F2F2}\textit{}} & \multicolumn{1}{l}{\cellcolor[HTML]{F2F2F2}\textit{}} & \multicolumn{1}{l}{\cellcolor[HTML]{F2F2F2}\textit{}} & \multicolumn{1}{l}{\cellcolor[HTML]{F2F2F2}\textit{}} & \multicolumn{1}{l}{\cellcolor[HTML]{F2F2F2}\textit{}} \\
 & {\ul 85.1} & \textbf{76.8} & 23.5 & {\ul 40.9} & 16.7 & 16.7 & 20.0 & 40.0 \\
\multirow{-2}{*}{Ada-R1} & \cellcolor[HTML]{DDEBF7}(324) & \cellcolor[HTML]{DDEBF7}(887) & \cellcolor[HTML]{DDEBF7}(3021) & \cellcolor[HTML]{DDEBF7}(2398) & \cellcolor[HTML]{DDEBF7}(3931) & \cellcolor[HTML]{DDEBF7}(4707) & \cellcolor[HTML]{DDEBF7}(3655) & \cellcolor[HTML]{DDEBF7}(2703) \\
 & 84.3 & 76.0 & 28.7 & 38.5 & 14.4 & \textbf{22.2} & 27.8 & 41.7 \\
\multirow{-2}{*}{HAPO} & \cellcolor[HTML]{DDEBF7}(918) & \cellcolor[HTML]{DDEBF7}(1920) & \cellcolor[HTML]{DDEBF7}(2125) & \cellcolor[HTML]{DDEBF7}(2975) & \cellcolor[HTML]{DDEBF7}(3888) & \cellcolor[HTML]{DDEBF7}(3753) & \cellcolor[HTML]{DDEBF7}(3118) & \cellcolor[HTML]{DDEBF7}(2671) \\
 & 84.9 & 74.9 & 35.7 & 38.2 & {\ul 21.1} & \textbf{22.2} & {\ul 28.6} & {\ul 43.7} \\
\multirow{-2}{*}{Query-Opt} & \cellcolor[HTML]{DDEBF7}(850) & \cellcolor[HTML]{DDEBF7}(1796) & \cellcolor[HTML]{DDEBF7}(2130) & \cellcolor[HTML]{DDEBF7}(2924) & \cellcolor[HTML]{DDEBF7}(3527) & \cellcolor[HTML]{DDEBF7}(3753) & \cellcolor[HTML]{DDEBF7}(3127) & \cellcolor[HTML]{DDEBF7}(2587) \\
 & 82.7 & 75.7 & 33.8 & 40.3 & 18.9 & 16.7 & 27.3 & 42.2 \\
\multirow{-2}{*}{LC-R1} & \cellcolor[HTML]{DDEBF7}(841) & \cellcolor[HTML]{DDEBF7}(2233) & \cellcolor[HTML]{DDEBF7}(1986) & \cellcolor[HTML]{DDEBF7}(2727) & \cellcolor[HTML]{DDEBF7}(3448) & \cellcolor[HTML]{DDEBF7}(3495) & \cellcolor[HTML]{DDEBF7}(3046) & \cellcolor[HTML]{DDEBF7}(2539) \\ \hline
\rowcolor[HTML]{F2F2F2} 
\textit{Arithmetic Merging} & \multicolumn{1}{l}{\cellcolor[HTML]{F2F2F2}\textit{}} & \multicolumn{1}{l}{\cellcolor[HTML]{F2F2F2}\textit{}} & \multicolumn{1}{l}{\cellcolor[HTML]{F2F2F2}\textit{}} & \multicolumn{1}{l}{\cellcolor[HTML]{F2F2F2}\textit{}} & \multicolumn{1}{l}{\cellcolor[HTML]{F2F2F2}\textit{}} & \multicolumn{1}{l}{\cellcolor[HTML]{F2F2F2}\textit{}} & \multicolumn{1}{l}{\cellcolor[HTML]{F2F2F2}\textit{}} & \multicolumn{1}{l}{\cellcolor[HTML]{F2F2F2}\textit{}} \\
 & 84.5 & 76.2 & 34.6 & 37.0 & 17.8 & 20.0 & 27.3 & 42.5 \\
\multirow{-2}{*}{Average Merging} & \cellcolor[HTML]{DDEBF7}(974) & \cellcolor[HTML]{DDEBF7}(1998) & \cellcolor[HTML]{DDEBF7}(2076) & \cellcolor[HTML]{DDEBF7}(2833) & \cellcolor[HTML]{DDEBF7}(3692) & \cellcolor[HTML]{DDEBF7}(3591) & \cellcolor[HTML]{DDEBF7}(2746) & \cellcolor[HTML]{DDEBF7}(2559) \\
 & 77.9 & 67.6 & 30.9 & 32.7 & 13.3 & 10.0 & 19.7 & 36.0 \\
\multirow{-2}{*}{Task Arithmetic} & \cellcolor[HTML]{DDEBF7}(926) & \cellcolor[HTML]{DDEBF7}(2872) & \cellcolor[HTML]{DDEBF7}(1127) & \cellcolor[HTML]{DDEBF7}(3692) & \cellcolor[HTML]{DDEBF7}(4097) & \cellcolor[HTML]{DDEBF7}(4019) & \cellcolor[HTML]{DDEBF7}(3332) & \cellcolor[HTML]{DDEBF7}(2867) \\
 & 78.4 & 72.6 & 33.8 & 36.0 & 10.0 & 13.3 & 25.8 & 38.6 \\
\multirow{-2}{*}{TIES Merging} & \cellcolor[HTML]{DDEBF7}(1134) & \cellcolor[HTML]{DDEBF7}(2523) & \cellcolor[HTML]{DDEBF7}(2751) & \cellcolor[HTML]{DDEBF7}(3353) & \cellcolor[HTML]{DDEBF7}(3924) & \cellcolor[HTML]{DDEBF7}(3865) & \cellcolor[HTML]{DDEBF7}(3055) & \cellcolor[HTML]{DDEBF7}(2944) \\
 & 67.8 & 56.2 & 20.2 & 23.6 & 3.3 & 10.0 & 20.2 & 28.8 \\
\multirow{-2}{*}{DARE} & \cellcolor[HTML]{DDEBF7}(734) & \cellcolor[HTML]{DDEBF7}(2005) & \cellcolor[HTML]{DDEBF7}(2355) & \cellcolor[HTML]{DDEBF7}(3055) & \cellcolor[HTML]{DDEBF7}(3935) & \cellcolor[HTML]{DDEBF7}(3804) & \cellcolor[HTML]{DDEBF7}(2622) & \cellcolor[HTML]{DDEBF7}(2644) \\ \hline
\rowcolor[HTML]{F2F2F2} 
\textit{Activation-based Merging} & \multicolumn{1}{l}{\cellcolor[HTML]{F2F2F2}\textit{}} & \multicolumn{1}{l}{\cellcolor[HTML]{F2F2F2}\textit{}} & \multicolumn{1}{l}{\cellcolor[HTML]{F2F2F2}\textit{}} & \multicolumn{1}{l}{\cellcolor[HTML]{F2F2F2}\textit{}} & \multicolumn{1}{l}{\cellcolor[HTML]{F2F2F2}\textit{}} & \multicolumn{1}{l}{\cellcolor[HTML]{F2F2F2}\textit{}} & \multicolumn{1}{l}{\cellcolor[HTML]{F2F2F2}\textit{}} & \multicolumn{1}{l}{\cellcolor[HTML]{F2F2F2}\textit{}} \\
 & 71.3 & 63.8 & 24.8 & 28.9 & 13.3 & 10.0 & 26.8 & 34.1 \\
\multirow{-2}{*}{Sens Merging} & \cellcolor[HTML]{DDEBF7}(720) & \cellcolor[HTML]{DDEBF7}(1611) & \cellcolor[HTML]{DDEBF7}(2774) & \cellcolor[HTML]{DDEBF7}(2684) & \cellcolor[HTML]{DDEBF7}(3547) & \cellcolor[HTML]{DDEBF7}(3467) & \cellcolor[HTML]{DDEBF7}(2950) & \cellcolor[HTML]{DDEBF7}(2536) \\
 & 84.9 & 75.8 & {\ul 35.7} & 40.1 & 16.7 & 20.0 & 25.8 & 42.7 \\
\multirow{-2}{*}{AIM} & \cellcolor[HTML]{DDEBF7}(808) & \cellcolor[HTML]{DDEBF7}(2405) & \cellcolor[HTML]{DDEBF7}(2470) & \cellcolor[HTML]{DDEBF7}(4271) & \cellcolor[HTML]{DDEBF7}(4031) & \cellcolor[HTML]{DDEBF7}(3600) & \cellcolor[HTML]{DDEBF7}(3472) & \cellcolor[HTML]{DDEBF7}(3008) \\
 & 78.4 & 71.4 & 28.7 & 33.8 & {\ul 21.1} & 18.9 & 27.8 & 40.0 \\
\multirow{-2}{*}{ACM} & \cellcolor[HTML]{DDEBF7}(962) & \cellcolor[HTML]{DDEBF7}(1236) & \cellcolor[HTML]{DDEBF7}(1350) & \cellcolor[HTML]{DDEBF7}(1850) & \cellcolor[HTML]{DDEBF7}(2689) & \cellcolor[HTML]{DDEBF7}(5535) & \cellcolor[HTML]{DDEBF7}(3815) & \cellcolor[HTML]{DDEBF7}(2491) \\ \hline
 & \textbf{85.2} & {\ul 76.3} & \textbf{36.0} & \textbf{41.2} & \textbf{26.7} & 20.0 & \textbf{28.8} & \textbf{44.9} \\
\multirow{-2}{*}{\modelname} & \cellcolor[HTML]{DDEBF7}(722) & \cellcolor[HTML]{DDEBF7}(1761) & \cellcolor[HTML]{DDEBF7}(1861) & \cellcolor[HTML]{DDEBF7}(2895) & \cellcolor[HTML]{DDEBF7}(4018) & \cellcolor[HTML]{DDEBF7}(3259) & \cellcolor[HTML]{DDEBF7}(2314) & \cellcolor[HTML]{DDEBF7}{\color[HTML]{00B050} \textbf{(2404)}} \\ \hline
\end{tabular}%
}
\label{tab:main_1_5B}
\end{table*}

\begin{table*}[ht!]
\caption{Evaluations of different methods on Qwen3-4B series models.}
\resizebox{0.9\textwidth}{!}{%
\begin{tabular}{ccccccccc}
\hline
\textbf{\diagbox{Model}{Bench}} & {\color[HTML]{1F2328} \textbf{GSM8K}} & {\color[HTML]{1F2328} \textbf{MATH500}} & {\color[HTML]{1F2328} \textbf{\begin{tabular}[c]{@{}c@{}}Minerva \\      Math\end{tabular}}} & \textbf{\begin{tabular}[c]{@{}c@{}}Olympiad\\      Bench\end{tabular}} & {\color[HTML]{1F2328} \textbf{AIME24}} & {\color[HTML]{1F2328} \textbf{AIME25}} & {\color[HTML]{1F2328} \textbf{GPQA}} & \textbf{Avg.} \\ \hline
 & 95.2 & 96.0 & 60.3 & 72.9 & 83.1 & 80.3 & 67.7 & 79.3 \\
\multirow{-2}{*}{\begin{tabular}[c]{@{}c@{}}Qwen3-4B-Thinking\\ (Long-CoT)\end{tabular}} & \cellcolor[HTML]{DDEBF7}(1521) & \cellcolor[HTML]{DDEBF7}(6125) & \cellcolor[HTML]{DDEBF7}(5501) & \cellcolor[HTML]{DDEBF7}(14405) & \cellcolor[HTML]{DDEBF7}(20469) & \cellcolor[HTML]{DDEBF7}(23912) & \cellcolor[HTML]{DDEBF7}(9032) & \cellcolor[HTML]{DDEBF7}(11566) \\ \hline
 & 94.5 & 93.0 & 43.0 & 62.8 & 66.7 & 50.0 & 54.5 & 66.4 \\
\multirow{-2}{*}{\begin{tabular}[c]{@{}c@{}}Qwen3-4B-Instruct\\ (Short-CoT)\end{tabular}} & \cellcolor[HTML]{DDEBF7}(374) & \cellcolor[HTML]{DDEBF7}(1670) & \cellcolor[HTML]{DDEBF7}(1446) & \cellcolor[HTML]{DDEBF7}(4199) & \cellcolor[HTML]{DDEBF7}(7046) & \cellcolor[HTML]{DDEBF7}(7368) & \cellcolor[HTML]{DDEBF7}(5086) & \cellcolor[HTML]{DDEBF7}(3884) \\ \hline
\rowcolor[HTML]{F2F2F2} 
\textit{Prompt-guided} & \multicolumn{1}{l}{\cellcolor[HTML]{F2F2F2}\textit{}} & \multicolumn{1}{l}{\cellcolor[HTML]{F2F2F2}\textit{}} & \multicolumn{1}{l}{\cellcolor[HTML]{F2F2F2}\textit{}} & \multicolumn{1}{l}{\cellcolor[HTML]{F2F2F2}\textit{}} & \multicolumn{1}{l}{\cellcolor[HTML]{F2F2F2}\textit{}} & \multicolumn{1}{l}{\cellcolor[HTML]{F2F2F2}\textit{}} & \multicolumn{1}{l}{\cellcolor[HTML]{F2F2F2}\textit{}} & \multicolumn{1}{l}{\cellcolor[HTML]{F2F2F2}\textit{}} \\
 & {\ul 95.6} & 94.8 & 57.7 & 56.3 & 66.7 & 63.3 & 65.2 & 71.4 \\
\multirow{-2}{*}{CoD} & \cellcolor[HTML]{DDEBF7}(1017) & \cellcolor[HTML]{DDEBF7}(3106) & \cellcolor[HTML]{DDEBF7}(3890) & \cellcolor[HTML]{DDEBF7}(5887) & \cellcolor[HTML]{DDEBF7}(11144) & \cellcolor[HTML]{DDEBF7}(12728) & \cellcolor[HTML]{DDEBF7}(6733) & \cellcolor[HTML]{DDEBF7}(6358) \\ \hline
\rowcolor[HTML]{F2F2F2} 
\textit{Training-based} & \multicolumn{1}{l}{\cellcolor[HTML]{F2F2F2}\textit{}} & \multicolumn{1}{l}{\cellcolor[HTML]{F2F2F2}\textit{}} & \multicolumn{1}{l}{\cellcolor[HTML]{F2F2F2}\textit{}} & \multicolumn{1}{l}{\cellcolor[HTML]{F2F2F2}\textit{}} & \multicolumn{1}{l}{\cellcolor[HTML]{F2F2F2}\textit{}} & \multicolumn{1}{l}{\cellcolor[HTML]{F2F2F2}\textit{}} & \multicolumn{1}{l}{\cellcolor[HTML]{F2F2F2}\textit{}} & \multicolumn{1}{l}{\cellcolor[HTML]{F2F2F2}\textit{}} \\
 & 95.3 & \textbf{96.4} & 58.5 & {\ul 71.4} & 72.2 & \textbf{68.9} & 65.7 & {\ul 75.5} \\
\multirow{-2}{*}{Ada-R1} & \cellcolor[HTML]{DDEBF7}(1161) & \cellcolor[HTML]{DDEBF7}(3746) & \cellcolor[HTML]{DDEBF7}(4115) & \cellcolor[HTML]{DDEBF7}(8455) & \cellcolor[HTML]{DDEBF7}(11324) & \cellcolor[HTML]{DDEBF7}(11969) & \cellcolor[HTML]{DDEBF7}(6395) & \cellcolor[HTML]{DDEBF7}(6738) \\ \hline
\rowcolor[HTML]{F2F2F2} 
\textit{Arithmetic Merging} & \multicolumn{1}{l}{\cellcolor[HTML]{F2F2F2}\textit{}} & \multicolumn{1}{l}{\cellcolor[HTML]{F2F2F2}\textit{}} & \multicolumn{1}{l}{\cellcolor[HTML]{F2F2F2}\textit{}} & \multicolumn{1}{l}{\cellcolor[HTML]{F2F2F2}\textit{}} & \multicolumn{1}{l}{\cellcolor[HTML]{F2F2F2}\textit{}} & \multicolumn{1}{l}{\cellcolor[HTML]{F2F2F2}\textit{}} & \multicolumn{1}{l}{\cellcolor[HTML]{F2F2F2}\textit{}} & \multicolumn{1}{l}{\cellcolor[HTML]{F2F2F2}\textit{}} \\
 & 95.4 & 95.4 & 57.7 & 70.5 & 73.3 & 57.8 & 64.7 & 73.5 \\
\multirow{-2}{*}{Average Merging} & \cellcolor[HTML]{DDEBF7}(1088) & \cellcolor[HTML]{DDEBF7}(3422) & \cellcolor[HTML]{DDEBF7}(4104) & \cellcolor[HTML]{DDEBF7}(7293) & \cellcolor[HTML]{DDEBF7}(9867) & \cellcolor[HTML]{DDEBF7}(10099) & \cellcolor[HTML]{DDEBF7}(7680) & \cellcolor[HTML]{DDEBF7}(6222) \\
 & 95.2 & 96.2 & 57.4 & 71.3 & 72.2 & {\ul 67.8} & 62.1 & 74.6 \\
\multirow{-2}{*}{Task Arithmetic} & \cellcolor[HTML]{DDEBF7}(1181) & \cellcolor[HTML]{DDEBF7}(3720) & \cellcolor[HTML]{DDEBF7}(3816) & \cellcolor[HTML]{DDEBF7}(7841) & \cellcolor[HTML]{DDEBF7}(10935) & \cellcolor[HTML]{DDEBF7}(11395) & \cellcolor[HTML]{DDEBF7}(6191) & \cellcolor[HTML]{DDEBF7}(6440) \\
 & 94.8 & 95.2 & {\ul \textbf{58.8}} & 70.2 & \textbf{75.6} & 60.0 & 63.6 & 74.0 \\
\multirow{-2}{*}{TIES Merging} & \cellcolor[HTML]{DDEBF7}(1170) & \cellcolor[HTML]{DDEBF7}(3411) & \cellcolor[HTML]{DDEBF7}(3782) & \cellcolor[HTML]{DDEBF7}(7168) & \cellcolor[HTML]{DDEBF7}(9735) & \cellcolor[HTML]{DDEBF7}(10891) & \cellcolor[HTML]{DDEBF7}(6598) & \cellcolor[HTML]{DDEBF7}(6108) \\
 & 95.1 & 94.2 & 57.0 & 67.7 & 64.4 & 56.7 & 58.6 & 70.5 \\
\multirow{-2}{*}{DARE} & \cellcolor[HTML]{DDEBF7}(1641) & \cellcolor[HTML]{DDEBF7}(3858) & \cellcolor[HTML]{DDEBF7}(4559) & \cellcolor[HTML]{DDEBF7}(7583) & \cellcolor[HTML]{DDEBF7}(11677) & \cellcolor[HTML]{DDEBF7}(12247) & \cellcolor[HTML]{DDEBF7}(6404) & \cellcolor[HTML]{DDEBF7}(6853) \\ \hline
\rowcolor[HTML]{F2F2F2} 
\textit{Activation-based Merging} & \multicolumn{1}{l}{\cellcolor[HTML]{F2F2F2}\textit{}} & \multicolumn{1}{l}{\cellcolor[HTML]{F2F2F2}\textit{}} & \multicolumn{1}{l}{\cellcolor[HTML]{F2F2F2}\textit{}} & \multicolumn{1}{l}{\cellcolor[HTML]{F2F2F2}\textit{}} & \multicolumn{1}{l}{\cellcolor[HTML]{F2F2F2}\textit{}} & \multicolumn{1}{l}{\cellcolor[HTML]{F2F2F2}\textit{}} & \multicolumn{1}{l}{\cellcolor[HTML]{F2F2F2}\textit{}} & \multicolumn{1}{l}{\cellcolor[HTML]{F2F2F2}\textit{}} \\
 & 95.4 & 95.6 & 58.1 & 68.2 & 64.4 & 62.2 & {\ul 67.7} & 73.1 \\
\multirow{-2}{*}{Sens Merging} & \cellcolor[HTML]{DDEBF7}(1271) & \cellcolor[HTML]{DDEBF7}(4111) & \cellcolor[HTML]{DDEBF7}(4349) & \cellcolor[HTML]{DDEBF7}(8564) & \cellcolor[HTML]{DDEBF7}(11640) & \cellcolor[HTML]{DDEBF7}(11985) & \cellcolor[HTML]{DDEBF7}(7826) & \cellcolor[HTML]{DDEBF7}(7106) \\
 & 95.1 & 94.8 & 55.9 & 70.5 & 64.4 & 60.0 & \textbf{70.7} & 73.1 \\
\multirow{-2}{*}{AIM} & \cellcolor[HTML]{DDEBF7}(1060) & \cellcolor[HTML]{DDEBF7}(2923) & \cellcolor[HTML]{DDEBF7}(3874) & \cellcolor[HTML]{DDEBF7}(6566) & \cellcolor[HTML]{DDEBF7}(10354) & \cellcolor[HTML]{DDEBF7}(9934) & \cellcolor[HTML]{DDEBF7}(8007) & \cellcolor[HTML]{DDEBF7}(6103) \\
 & 94.7 & 94.0 & 56.2 & 68.4 & 70.0 & 54.4 & 61.6 & 71.3 \\
\multirow{-2}{*}{ACM} & \cellcolor[HTML]{DDEBF7}(925) & \cellcolor[HTML]{DDEBF7}(3208) & \cellcolor[HTML]{DDEBF7}(3197) & \cellcolor[HTML]{DDEBF7}(6891) & \cellcolor[HTML]{DDEBF7}(10988) & \cellcolor[HTML]{DDEBF7}(11080) & \cellcolor[HTML]{DDEBF7}(6141) & \cellcolor[HTML]{DDEBF7}(6061) \\ \hline
 & \textbf{95.7} & \textbf{96.4} & \textbf{60.3} & \textbf{72.0} & \textbf{75.6} & {\ul 67.8} & 66.7 & \textbf{76.4} \\
\multirow{-2}{*}{\modelname} & \cellcolor[HTML]{DDEBF7}(1086) & \cellcolor[HTML]{DDEBF7}(3191) & \cellcolor[HTML]{DDEBF7}(4051) & \cellcolor[HTML]{DDEBF7}(7225) & \cellcolor[HTML]{DDEBF7}(8394) & \cellcolor[HTML]{DDEBF7}(10157) & \cellcolor[HTML]{DDEBF7}(7612) & \cellcolor[HTML]{DDEBF7}{\color[HTML]{00B050} \textbf{(5959)}} \\ \hline
\end{tabular}%
}
\label{tab:main_4B}
\end{table*}
In this section, we conduct comprehensive experiments to evaluate the effectiveness of \modelname on seven reasoning benchmarks and across two model scales. 
Specifically, our investigation aims to  answer the following research questions:
\begin{itemize}[leftmargin=*]
    \item \textbf{RQ1}: How effective is \modelname for adaptive reasoning?
    \item \textbf{RQ2}: Is each key component in \modelname effective?
    \item \textbf{RQ3}: How do the hyper-parameters influence \modelname?
    \item \textbf{RQ4}: How run-time efficient is our method compared with other adaptive reasoning methods?
\end{itemize}

\subsection{Experimental Setup}
\label{sec:exp_setup}
\subsubsection{\textbf{Datasets and Evaluation Protocols.}}
We evaluate \modelname on seven widely used reasoning benchmarks. Following prior practice~\citep{Ada-R1-neurips25-luo2025dpoMerge,ACM-merging-neurips25}, we treat GSM8K~\citep{GSM8K-dataset} test set, MATH500~\citep{hendrycks2021-Math500-dataset} test set, and AIME24~\citep{aime24_dataset} as in-distribution (ID) evaluation data. 
We further assess its generalization performance on out-of-distribution (OOD) benchmarks, including AIME25~\citep{aime25_dataset}, Minerva Math~\citep{SolveReasonLLM-nips22-Aitor}, OlympiadBench~\citep{he-etal-2024-olympiadbench}, and GPQA~\citep{rein2024-gpqa-dataset}.
These datasets are summarized below:
\begin{itemize}[leftmargin=*]
    \item GSM8K(1319): Grade-school and middle-school word problems emphasizing Short-CoT arithmetic reasoning.
    \item MATH500(500): Comprises 500 problems spanning arithmetic, algebra, geometry, and calculus, with five difficulty levels.
    \item Minerva Math(272): consists of 272 undergraduate-level STEM problems harvested from MIT’s OpenCourseWare.
    \item OlympiadBench(675): A bilingual math+physics Olympiad-style benchmark sourced from olympiads and Gaokao.
    \item AIME24(30): An Olympiad-style set assessing logical deduction and advanced problem-solving skills.
    \item AIME25(30): An updated benchmark following AIME24.
    \item GPQA(198): A challenging graduate-level subset with multiple-choice questions authored by domain experts in biology, physics, and chemistry.
\end{itemize}

We report both \textbf{accuracy} and \textbf{response length} (average number of generated tokens) on each benchmark to jointly characterize solution quality and inference cost. In addition, we also report the average accuracy and the average response length across all benchmarks. All evaluations are conducted with the AReal evaluation framework~\citep{AReal-evaluation-fu2025areal}. For each benchmark, we independently execute the model five times with different random seeds and report the averaged results, and all results are statistically significant at $p <$ 0.05.

\subsubsection{\textbf{Implementation Details.}}
To examine our method's generalization ability across model families and scales, we consider two Long/Short-CoT pairs:
(i) \textit{DeepSeek-R1-Distill-Qwen-1.5B}~\citep{deepseekai2025deepseek-r1-qwen25} (Long-CoT) and \textit{Qwen2.5-Math-1.5B}~\citep{yang2024qwen25-math} (Short-CoT);
(ii) \textit{Qwen3-4B-Thinking} (Long-CoT)~\citep{qwen3technicalreport} and \textit{Qwen3-4B-Instruct}~\citep{qwen3technicalreport} (Short-CoT).
For pattern-labeled dataset construction, we randomly sample a total of 128 questions from the GSM8K, MATH500, and AIME24, and draw \(k{=}12\) responses per question from each of the two base models. 
By default, we calibrate the merging coefficient of the top 30\% layers and correct the hidden state of the remaining layers with $p_S=0.3$.
We select the learning rate and the number of training epochs via grid search over \(\{0.1, 0.01, 0.001\}\) and \(\{50, 100\}\), respectively. All experiments are conducted on a single NVIDIA RTX A5000 GPU.

\subsubsection{\textbf{Baselines.}}
To comprehensively evaluate the effectiveness of \modelname, we compare it with five categories of baselines: (1) Long- and Short-CoT base models; (2) the prompt-guided Chain-of-Draft (CoD)~\citep{xu2025-CoD-ChainofDraft-prompt}; (3) training-based methods, including Ada-R1~\citep{Ada-R1-neurips25-luo2025dpoMerge}, HAPO~\citep{huang2025-HAPO-RLbase}, Query-Opt~\citep{arora_Zanette2025-RLbase-reasoning}, and LC-R1~\citep{cheng2025-LC-R1-RLbase}; (4) arithmetic merging methods, including Average Merging~\citep{ModelSoups-averageWeight-2022Kamalika}, Task Arithmetic~\citep{editing-TAMerge-ilharco2023}, TIES~\citep{TIES-merging-NEURIPS2023}, and DARE~\citep{DARE-merging-SuperMarioICML24}; and (5) activation-based merging methods, including Sens Merging~\citep{liu2025sens-merging}, AIM~\citep{AIM-merging-nobari2025}, and ACM~\citep{ACM-merging-neurips25}.

\subsection{Main Results (RQ1)}
For the two model pairs, we report the results in Tables~\ref{tab:main_1_5B} and~\ref{tab:main_4B}, respectively. Overall, the results show that \modelname consistently achieves a strong accuracy--efficiency trade-off across both model series. The Long-CoT models provide strong reasoning performance, but at the cost of longer responses. 
In contrast, the Short-CoT models generate much shorter responses but suffer from clear accuracy degradation, especially on challenging reasoning benchmarks such as AIME24 and AIME25. 

\subsubsection{\textbf{Comparison with Prompt-guided and Merging-based Methods.}}
We first compare \modelname with prompt-guided and merging baselines, which do not require adaptive reasoning training. 
From the experimental results in the 1.5B setting (Table~\ref{tab:main_1_5B}), we observe that the prompt-guided method CoD improves efficiency over the Long-CoT model, but still underperforms it in average accuracy. 
Arithmetic merging methods reduce generation length considerably, yet incur noticeable accuracy degradation.
Activation-based merging baselines provide more competitive trade-offs, with AIM reaching 42.7 average accuracy and ACM reaching 40.0 average accuracy. 
However, these methods still underperform \modelname. 
\modelname achieves the highest average accuracy of 44.9 while using only 2404 tokens on average, reducing the response length by 70.5\% relative to the Long-CoT model and improving average accuracy by 0.6 percentage points. 

Similar trends are observed in the 4B setting in Table~\ref{tab:main_4B}. 
By comparison, \modelname reaches 76.4 average accuracy with 5959 tokens on average. 
This corresponds to a 48.5\% reduction in response length relative to the Qwen3 Long-CoT model, while maintaining stronger accuracy than all training-free and merging baselines. 
At the benchmark level, \modelname achieves the best performance on GSM8K and MATH500, matches the Long-CoT model on Minerva Math, and remains competitive on more difficult benchmarks such as OlympiadBench, AIME24, and AIME25.

\subsubsection{\textbf{Comparison with Training-based Methods.}}
We further compare \modelname with training-based adaptive reasoning methods in the Qwen2.5-1.5B setting. 
These methods post-train the model toward adaptive or concise reasoning through reinforcement learning. 
They generally obtain competitive accuracy--length trade-offs, with Query-Opt being the strongest baseline in this group, reaching 43.7 average accuracy with an average response length of 2587 tokens.
In comparison, \modelname improves the average accuracy from 43.7 to 44.9 while further reducing the average response length from 2587 to 2404 tokens. 
This result indicates that lightweight query-adaptive merging can match or surpass training-based adaptive reasoning methods under the same Qwen2.5-1.5B evaluation setting, without relying on expensive post-training.

For the Qwen3-4B setting, we include Ada-R1 as a representative training-based baseline. 
We do not additionally reproduce GRPO-based or GRPO-variant methods, including HAPO, Query-Opt, and LC-R1, on the 4B model. 
In our experimental environment, applying RL-based optimization to a 4B model requires high GPU memory and repeated rollout generation, which exceeds our available computational budget. 
Therefore, our main comparison with these RL-based methods is conducted on the Qwen2.5-1.5B model, where their training cost is more tractable.

In summary, \modelname consistently reduces inference cost while preserving strong reasoning performance. 
Across both model series, it outperforms prompt-guided and merging-based baselines, and in the Qwen2.5-1.5B setting, it also achieves a more favorable trade-off than training-based adaptive reasoning methods. 
These results support \modelname as an efficient alternative to costly adaptive reasoning training.

\subsection{Ablation Study  (RQ2)}
Table~\ref{tab:ablation} presents the ablation results for the key components of \modelname. 
The full model consistently provides the most favorable balance between reasoning quality and generation efficiency, demonstrating that its improvements do not arise from a single isolated design choice but from the coordinated effect of all components. 
Removing the layer selective strategy (w/o LS) weakens performance most noticeably on more challenging reasoning tasks, suggesting that reasoning divergence are concentrated in high-divergence layers and should not be treated uniformly across all layers. 
Removing the feature-alignment calibration (w/o Cal.) further degrades the merged model, indicating that simple parameter interpolation is insufficient to recover query-adaptive Long-/Short-CoT behavior without explicitly aligning the merged representations with the preferred reasoning pattern. 
In addition, removing hidden-state correction (w/o Corr.) leads to less stable performance and longer generations, showing that even lower divergence layers can introduce residual hidden-state distortion after merging. 
Overall, the combination of layer selection, feature-alignment calibration, and hidden-state correction contributes to the success of \modelname in learning query-adaptive reasoning and achieving an effective balance between reasoning accuracy and generation efficiency.


\begin{table}[t]
\caption{Ablation study of each component.}
\resizebox{0.9\columnwidth}{!}{%
\begin{tabular}{ccccc}
\hline
\textbf{\diagbox{Model}{Bench}} & {\color[HTML]{1F2328} \textbf{GSM8K}} & {\color[HTML]{1F2328} \textbf{MATH500}} & {\color[HTML]{1F2328} \textbf{AIME24}} & \textbf{Avg.} \\ \hline
 & 79.0 & 80.6 & 29.2 & 62.9 \\
\multirow{-2}{*}{Long-CoT Model} & \cellcolor[HTML]{DDEBF7}(978) & \cellcolor[HTML]{DDEBF7}(3741) & \cellcolor[HTML]{DDEBF7}(17465) & \cellcolor[HTML]{DDEBF7}(7395) \\ \hline
 & 85.2 & 76.3 & 26.7 & 62.7 \\
\multirow{-2}{*}{\modelname} & \cellcolor[HTML]{DDEBF7}(722) & \cellcolor[HTML]{DDEBF7}(1761) & \cellcolor[HTML]{DDEBF7}(4018) & \cellcolor[HTML]{DDEBF7}(2167) \\ \hline
 & 85.1 & 74.2 & 16.7 & 58.7 \\
\multirow{-2}{*}{w/o LS} & \cellcolor[HTML]{DDEBF7}(801) & \cellcolor[HTML]{DDEBF7}(1856) & \cellcolor[HTML]{DDEBF7}(4481)& \cellcolor[HTML]{DDEBF7}(2379) \\ \hline
 & 84.1 & 73.3 & 10 & 55.8 \\
\multirow{-2}{*}{w/o Cal.} & \cellcolor[HTML]{DDEBF7}(731) & \cellcolor[HTML]{DDEBF7}(1734) & \cellcolor[HTML]{DDEBF7}(4241) & \cellcolor[HTML]{DDEBF7}(2235) \\ \hline
 & 84.5 & 75.1 & 20.0 & 59.9 \\
\multirow{-2}{*}{w/o Corr.} & \cellcolor[HTML]{DDEBF7}(838) & \cellcolor[HTML]{DDEBF7}(1941) & \cellcolor[HTML]{DDEBF7}(4317) & \cellcolor[HTML]{DDEBF7}(2365) \\ \hline
\end{tabular}%
}
\label{tab:ablation}
\end{table}
\begin{figure*}[pt]
  \includegraphics[width=0.9\textwidth]{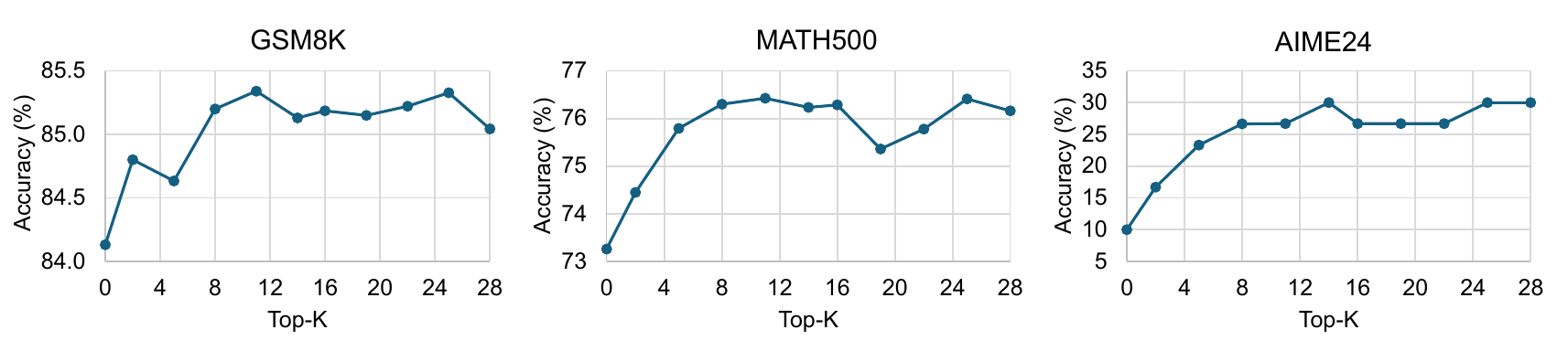}
  \caption{Influence of the number of selected layers $K$.}
  \label{fig:hyper-train-ratio}
\end{figure*}
\begin{figure*}[pt]
  \includegraphics[width=0.9\textwidth]{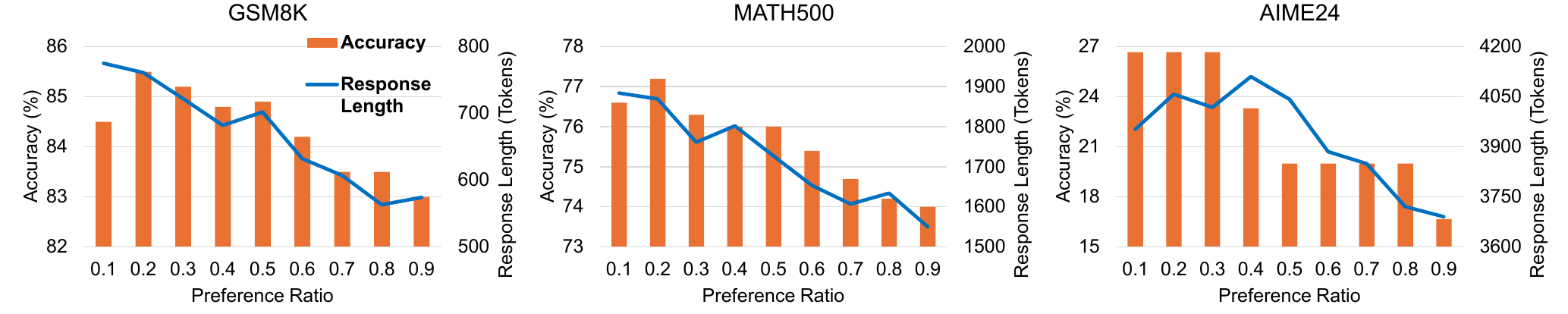}
  \caption{Influence of different preference ratio.}
  \label{fig:hyper-prefer-ratio}
\end{figure*}
\subsection{Hyper-parameter Analysis  (RQ3)}
We further analyze the sensitivity of \modelname to two key hyper-parameters: the number of selected layers (Top-$K$) for feature-alignment calibration and the preference ratio ($p_S$) used in hidden-state correction.

\noindent{\textbf{Effect of the number of selected layers.}}
Figure~\ref{fig:hyper-train-ratio} presents the influence of varying the value $K$ on GSM8K, MATH500, and AIME24. 
As the number of calibrated layers increases from zero to the top eight high-divergence layers, \modelname exhibits a rapid improvement across all benchmarks. 
This observation indicates that calibrating these layers is sufficient to recover a portion of the adaptive reasoning capability. 
When additional layers are included beyond the top eight, performance continues to improve marginally but remains broadly stable overall. 
This saturation pattern suggests that the dominant reasoning divergence is concentrated in a limited subset of layers, whereas lower divergence layers contribute comparatively minor incremental benefits. 
These findings provide support for the proposed layer selective strategy, which concentrates calibration on a subset of layers while avoiding the unnecessary cost of full-model calibration.

\noindent{\textbf{Effect of the preference ratio.}}
Figure~\ref{fig:hyper-prefer-ratio} reports the effect of the preference ratio on both accuracy and response length. 
As the preference ratio increasingly favors the concise reasoning pattern (Short-CoT), response length consistently decreases across three benchmarks, confirming that hidden-state correction provides an effective control for generation efficiency. 
The corresponding accuracy trend depends on task difficulty: GSM8K remains relatively robust under stronger concise preferences, MATH500 shows only mild degradation, while AIME24 exhibits a more significant accuracy drop when the model is overly biased toward Short-CoT behavior. 
This suggests that simple problems can be solved efficiently with concise reasoning, whereas complex math problems still require sufficient detailed reasoning chains.

\begin{table}[]
\caption{Training efficiency comparison on the Qwen2.5-1.5B setting. ``RT'' denotes the run-time for training or merging.}
\resizebox{\columnwidth}{!}{%
\begin{tabular}{lccccc}
\hline
 \textbf{Method} & \textbf{Paradigm} & \textbf{Avg. Acc.} & \textbf{Data Size} & \textbf{Trainable Params.} & \textbf{RT} \\ \hline
\textbf{Ada-R1} & DPO & 40.0 & 10863 & Full model & 8 hrs \\
\textbf{HAPO} & GRPO & 41.7 & 2000 & Full model & \textgreater{}24 hrs \\
\textbf{Query-Opt} & GRPO & 43.7 & 3200 & Full model & 20 hrs \\
\textbf{LC-R1} & GRPO & 42.2 & 2500 & Full model & \textgreater{}24 hrs \\
\textbf{AIM} & Merge & 42.7 & 1000 & 0 & 15.1  mins \\
\textbf{ACM} & Merge & 40.0 & 1000 & 0 & 13.6  mins \\
\textbf{\modelname} & Merge & 44.9 & 128 & \begin{tabular}[c]{@{}c@{}}Merging coeff.\\ in top-8 layers\end{tabular} & 20.7  mins \\ \hline
\end{tabular}%
}
\label{tab:train_efficiency}
\vspace{-0.6em}
\end{table}

\subsection{Run-time Efficiency Study (RQ4)}
Table~\ref{tab:train_efficiency} compares the training or merging efficiency of different adaptive reasoning methods in the Qwen2.5-1.5B setting.
All training-based approaches in this comparison require preference optimization or RL-based training, which involve rollout generation and full model optimization.
For these methods, the policy model repeatedly generates multiple reasoning trajectories for each query before reward-based policy updates.
In contrast, merging methods avoid full model training and provide a more efficient route to adaptive reasoning.
Among these methods, \modelname uses only a compact calibration set and calibrates the merging coefficients in a selected subset of layers, rather than updating all model parameters.
As a result, its runtime remains comparable to other merging-based baselines while delivering stronger reasoning performance.
Overall, these results demonstrate that \modelname provides an efficient and effective alternative to costly RL-based adaptive reasoning, achieving a favorable balance between optimization cost and reasoning quality.

\section{Related Work}
\subsection{Adaptive Reasoning}
Adaptive reasoning improves efficiency by dynamically selecting an appropriate reasoning pattern conditioned on query difficulty, avoiding overthinking on simple problems while retaining deeper deliberation for harder ones~\citep{sui2025stopoverthinkingsurveyefficient,zhu2025-Survey-AdaptiveThinkInLRM}. 
Existing approaches broadly fall into two lines: prompt-guided and training-based methods. 
\textbf{Prompt-guided} approaches~\citep{ConciseCoT-prompt-2024Renze,gong2025-CoUT,xu2025-CoD-ChainofDraft-prompt} exploit the instruction-following capability of LRMs by imposing explicit constraints through carefully crafted prompts to elicit concise reasoning chains. 
For instance, CoUT~\citep{gong2025-CoUT} first guides LRMs to internalize thinking processes and then generates the final answer using several token-efficient decoding strategies.
However, their effectiveness can be brittle, as it depends heavily on prompt design, constraint choices, and the model’s instruction-following behavior. 
\textbf{Training-based} methods explicitly optimize models to control or reduce response length, e.g., by fine-tuning LRMs on variable-length CoT supervision~\citep{Ada-R1-neurips25-luo2025dpoMerge,qiao-etal-emnlp2025-conciseSFTLRM} or by applying reinforcement learning with length-aware rewards~\citep{adaptthink-zhang-etal-2025,arora_Zanette2025-RLbase-reasoning}. 
For example, \citet{arora_Zanette2025-RLbase-reasoning} post-train LRMs with RL using a length-penalty reward that combines correctness and normalized length of sampled solutions, assigning higher reward scores to shorter correct solutions. 
While effective, these approaches typically require substantial training data and incur high computational cost.

\subsection{Model Merging}
Model merging~\citep{modelmerginginLlmsMllms-Yang2024} is an emerging technique that combines parameters from multiple pretrained or fine-tuned models with complementary capabilities into a single model, typically without access to the original training data and without expensive end-to-end retraining. 
We divide existing model merging methods into pre-merging and during-merging.
(i) Pre-merging methods focus on creating better conditions for merging. 
Recent works have investigated architectural transformation to convert heterogeneous models into homogeneous models~\citep{nguyen2023CLAFusion,xu2024train-free-heterogeneous-merge} and weight alignment to place models in the same basin~\citep{du2025adamms,wan2024knowledge-fusion}.
For instance, 
AdaMMS~\citep{du2025adamms} proposes a weight mapping function before performing merging on different backbones.
(ii) During-merging methods aim to design merging techniques on multiple models. 
Early work such as Model Soups~\citep{ModelSoups-averageWeight-2022Kamalika} demonstrates that simple weight averaging over multiple checkpoints can improve overall performance. 
Beyond averaging, more advanced methods, including TIES-Merging~\citep{TIES-merging-NEURIPS2023} and DARE~\citep{DARE-merging-SuperMarioICML24}, aim to reduce interference among task vectors by selectively retaining and combining only the most salient parameter updates. 

More recently, several studies~\citep{wu2025unlockingefficientL2S,ACM-merging-neurips25,revisitingMI-wu2025} have explored model merging for efficient reasoning by combining a slow-thinking Long-CoT model with a fast-thinking Short-CoT model. 
These approaches typically aim to produce a single merged model that globally shorten reasoning chains while retaining accuracy. 
In contrast, our work focuses on query-level adaptive reasoning: we calibrate the merging coefficients of a merged model to selectively follow Long-CoT or Short-CoT patterns depending on the input, thereby improving the accuracy and efficiency trade-off.

\section{Conclusion}
In this paper, we propose \modelname, a novel and efficient activation-based merging framework for query-adaptive reasoning.
\modelname constructs a compact pattern-labeled calibration dataset and designs a layer-wise merging framework, which calibrates high-divergence layers through feature alignment and contrastive shaping while applying hidden-state correction to the remaining layers.
The experimental results demonstrate that \modelname significantly reduces response length while preserving model performance, highlighting the effectiveness of query-adaptive model merging for adaptive reasoning.

\section*{GenAI Usage Disclosure}
ChatGPT-5.5 was used for language polishing purposes during the preparation of this manuscript. After using this tool, the authors reviewed and edited the content as needed and take full responsibility for the content of the paper.

\begin{acks}
The Australian Research Council partially supports this work under the streams of Future Fellowship (Grant No. FT210100624), the Discovery Project (Grant No. DP260100326), and the Linkage Project (Grant No. LP230200892, LP240200546 and LP250200778)
\end{acks}

\bibliographystyle{ACM-Reference-Format}
\balance
\bibliography{sample-base}

\end{document}